\documentclass[11pt]{article}

\usepackage[preprint]{acl}

\usepackage{times}
\usepackage{latexsym}
\usepackage{hyperref}
\usepackage{booktabs}
\usepackage[T1]{fontenc}
\usepackage[table]{xcolor}
\usepackage{comment}

\usepackage{caption}
\usepackage{subcaption}
\usepackage{minted}

\usepackage{listings}
\usepackage{xcolor}

\definecolor{lightgray}{gray}{0.95}

\lstdefinestyle{ragvue}{
    backgroundcolor=\color{lightgray},
    frame=single,
    framerule=0.5pt,
    basicstyle=\ttfamily\footnotesize,
    keywordstyle=\color{blue},
    commentstyle=\color{green!50!black},
    stringstyle=\color{purple},
    showstringspaces=false,
    breaklines=true
}

\usepackage[utf8]{inputenc}

\usepackage{microtype}
\usepackage{multirow}

\usepackage{inconsolata}

\usepackage{graphicx}
\usepackage{amsmath}

\title{\textsc{RAGVue}: A Diagnostic View for Explainable and Automated Evaluation of Retrieval-Augmented Generation}

\author{
  \textbf{Keerthana Murugaraj\textsuperscript{1}},
  \textbf{Salima Lamsiyah\textsuperscript{1}},
  \textbf{Martin Theobald\textsuperscript{1}} \\
  \textsuperscript{1}University of Luxembourg, Department of Computer Science (DCS), \\
  Faculty of Science, Technology and Medicine (FSTM), Esch-sur-Alzette, Luxembourg \\
   \small{
 \textbf{Correspondence:} \href{mailtos:keerthana.murugaraj@uni.lu}{keerthana.murugaraj@uni.lu}
 }
}

\begin{document}
\maketitle

\begin{abstract}
Evaluating Retrieval-Augmented Generation (RAG) systems remains a challenging task: existing metrics often collapse heterogeneous behaviors into single scores and provide little insight into whether errors arise from retrieval, reasoning, or grounding. In this paper, we introduce \textsc{RAGVue}, a diagnostic and explainable framework for automated, reference-free evaluation of RAG pipelines. \textsc{RAGVue} decomposes RAG behavior into retrieval quality, answer relevance and completeness, strict claim-level faithfulness, and judge calibration.  Each metric includes a structured explanation, making the evaluation process transparent. Our framework supports both manual metric selection and fully automated agentic evaluation. It also provides a Python API, CLI, and a local Streamlit interface for interactive usage. In comparative experiments, \textsc{RAGVue} surfaces fine-grained failures that existing tools such as RAGAS often overlook. We showcase the full \textsc{RAGVue} workflow and illustrate how it can be integrated into research pipelines and practical RAG development. The source code and detailed instructions on usage are publicly available on GitHub \footnote{\url{https://github.com/KeerthanaMurugaraj/RAGVue}}. 
\end{abstract}

\section{Introduction}

Retrieval-Augmented Generation (RAG) combines a pretrained (parametric) language model with an external retriever that supplies relevant documents at inference time~\cite{lewis2020retrieval,guu2020retrieval}. By conditioning generation on retrieved passages, RAG systems effectively tackle knowledge-intensive tasks while making their evidence explicit and easier to maintain than finetuning internal model weights~\cite{lewis2020retrieval,10.5555/3648699.3648950}. This paradigm has rapidly become a default solution for building search assistants, analytical tools, customer-support bots, and domain-specific copilots across high-stakes settings such as finance, healthcare, and law~\cite{song2024measuring,rosenthal-etal-2025-clapnq}. Recent benchmarks further stress-test RAG in multi-hop and multi-turn scenarios (e.g., StrategyQA~\cite{geva-etal-2021-aristotle}, mtRAG~\cite{10.1162/TACL.a.19}, CLAPnq~\cite{rosenthal-etal-2025-clapnq}), underscoring the need for robust and fine-grained evaluation of the full RAG pipeline. 

Evaluating RAG is harder than evaluating a standalone language model because errors can arise from retrieval (irrelevant or missing evidence), generation (off-topic, incomplete, or incoherent answers), or grounding (unsupported or contradictory claims despite retrieved context)~\cite{es2024ragas,saad-falcon-etal-2024-ares,ru2024ragchecker}. Recent surveys argue that global "end-to-end" scores obscure these components and advocate decomposing the evaluation into retrieval quality, answer quality, and evidence support~\cite{yu2024evaluation,gan2025retrieval}. They also emphasize a crucial distinction between \emph{faithfulness} to retrieved evidence and \emph{factual correctness} with respect to world knowledge: a response may be factually true but unsupported by its citations, or fully grounded in outdated or erroneous evidence~\cite{min-etal-2023-factscore,sorodoc-etal-2025-garage}. Temporal drift~\cite{ouyang-etal-2025-hoh}, unanswerability~\cite{peng-etal-2025-unanswerability}, and privacy or policy violations in retrieved content~\cite{zeng-etal-2025-rag,song2024measuring} introduce additional evaluation axes that simple accuracy-style metrics cannot capture.

Human annotation and gold references are expensive and brittle under domain shift~\cite{saad-falcon-etal-2024-ares,rosenthal-etal-2025-clapnq}, motivating reference-free \emph{LLM-as-a-judge} methods that are widely used in NLG evaluation~\cite{wang-etal-2023-chatgpt,kocmi-federmann-2023-large,10.5555/3666122.3668142}. Despite progress (e.g., G-Eval~\cite{liu-etal-2023-g}, AutoCalibrate~\cite{liu-etal-2024-calibrating}, SelfCheckGPT~\cite{manakul-etal-2023-selfcheckgpt}), LLM judges remain prompt-sensitive, unstable, and prone to self-preference bias~\cite{10.5555/3737916.3740113,schroeder2024can,liu-etal-2025-judge}. Existing RAG-focused evaluators, including RAGAS~\cite{es2024ragas}, ARES~\cite{saad-falcon-etal-2024-ares}, RAGChecker~\cite{ru2024ragchecker}, and RAG-Zeval~\cite{li-etal-2025-rag} have expanded coverage. However, two core gaps persist: metrics often collapse heterogeneous behaviors into non-diagnostic scalar scores, and grounding checks remain permissive, missing fine-grained factual errors~\cite{es2024ragas,niu2024ragtruth,song2024measuring}.

To address these limitations, we introduce \textsc{RAGVue}, a reference-free, explainable evaluation framework that offers \textit{diagnostic results} rather than purely numerical assessments. \textsc{RAGVue} decomposes RAG performance into retrieval quality, answer quality, and factual grounding~\cite{yu2024evaluation,gan2025retrieval}, enforcing \emph{strict faithfulness} by crediting only claim-level evidence explicitly supported in the retrieved context. This yields a more conservative alternative to semantic-inference metrics~\cite{es2024ragas,min-etal-2023-factscore,niu2024ragtruth,zhu-etal-2025-rageval}. \textsc{RAGVue} additionally introduces a \emph{judge-calibration} score quantifying agreement across LLM evaluators, making stability issues in LLM-as-a-judge setups explicit~\cite{liu-etal-2025-judge,schroeder2024can,10.5555/3737916.3740113}. The framework supports both \emph{manual} metric selection and an \emph{agentic} mode, in which an internal orchestrator automatically chooses and aggregates metrics. Moreover, we provide a Python API, command-line interface (CLI), and a Streamlit-based user interface (UI) for a seamless integration into research workflows. Finally, on a multihop StrategyQA-derived benchmark~\cite{geva-etal-2021-aristotle}, \textsc{RAGVue} reveals fine-grained failures that approaches based on scalar metrics, such as RAGAS~\cite{es2024ragas}, fail to capture.

\section{Related Work}

\paragraph{RAG \& Evaluation Challenges.} Retrieval-Augmented Generation (RAG) integrates external evidence into LLMs to reduce hallucinations and improve grounding~\cite{lewis2020retrieval,kocmi-federmann-2023-large,wang-etal-2023-chatgpt}. Early models such as REALM~\cite{guu2020retrieval} and RAG~\cite{lewis2020retrieval} showed strong gains on knowledge-intensive tasks, followed by advances in retrieval and generation (e.g., late-interaction retrievers~\cite{khattab2021relevance} and few-shot RAG tuning~\cite{10.5555/3648699.3648950}). However, the pipeline-based nature of RAG introduces unique evaluation challenges. Performance must be assessed across components, including retriever relevance and evidence coverage, generator quality, and grounding faithfulness~\cite{saad-falcon-etal-2024-ares}. Recent surveys argue that end-to-end scores obscure these dimensions and call for separate evaluation of retrieval quality and grounding fidelity~\cite{yu2024evaluation,gan2025retrieval}. Despite having access to documents, RAG models still tend to hallucinate or rely on outdated evidence, motivating benchmarks for temporal drift and attribution~\cite{ouyang-etal-2025-hoh}. Moreover, faithfulness and factual correctness may diverge: a response can be true but unsupported, or well-grounded yet incorrect~\cite{khattab2021relevance}.  We follow this line by separately evaluating retrieval, answer quality, and grounding with strict evidence criteria.

\paragraph{RAG Evaluation Frameworks \& Benchmarks.}
Recent work has introduced automatic evaluators for RAG. RAGAS~\cite{es2024ragas} provides reference-free metrics for context relevance, answer coherence, and coarse groundedness via static prompt-based queries. ARES~\cite{saad-falcon-etal-2024-ares} increases robustness by fine-tuning smaller LMs on human labels, offering explicit relevance and faithfulness scores with confidence estimates. RAGChecker~\cite{ru2024ragchecker} adds diagnostic checks for passage usage and claim-level grounding. Benchmarks such as RAGTruth~\cite{niu2024ragtruth} and MEMERAG~\cite{cruz-blandon-etal-2025-memerag} target hallucinations and multilingual settings, while HoH~\cite{ouyang-etal-2025-hoh} and Unanswerability-Eval~\cite{peng-etal-2025-unanswerability} test temporal drift and unanswerable queries. Furthermore, LLM-as-a-judge approaches are widely adopted~\cite{wang-etal-2023-chatgpt,10.5555/3666122.3668142}, including G-Eval~\cite{liu-etal-2023-g}, AutoCalibrate~\cite{liu-etal-2024-calibrating}, and FactScore~\cite{min-etal-2023-factscore}, but remain prompt-sensitive and biased toward model families~\cite{10.5555/3737916.3740113,liu-etal-2025-judge}. More reliable methods such as JudgeLM~\cite{liu-etal-2025-judge} and RAG-Zeval~\cite{li-etal-2025-rag} seek stability through consensus and reasoning-based ranking. Building on these insights, \textsc{RAGVue} uses reference-free LLM judges while addressing key limitations by decomposing scores (retrieval vs.\ coverage; answer relevance vs.\ completeness), enforcing strict claim-level faithfulness, and providing fine-grained explanations and stability checks. It also includes an \emph{agentic} evaluation mode that automatically selects and orchestrates metrics, producing structured summaries ready for debugging and comparison.

\section{\textsc{RAGVue} Framework Overview}
This section introduces our \textsc{RAGVue} evaluation framework. We first outline its core metrics, then describe its two operational modes, and conclude with its programmatic and interactive UIs.

\subsection{\textsc{RAGVue} Metrics} \label{RAGVue_metrics}
We describe seven \textsc{RAGVue} metrics across three dimensions: (1) retrieval, (2) answer quality, and (3) grounding and stability, with a summary provided in Appendix~\ref{app:summary_metrics} (Table~\ref{tab:RAGVue-metrics-summary}).

\subsubsection{Retrieval Relevance}\label{RR}
This metric measures whether the retrieved contexts (C) are actually useful for answering the question (Q). For each context chunk, an LLM judge assigns a relevance score $r_i$ in $[0,1]$ using a predefined range\footnote{Default ranges: 1.0-0.9 for direct answer-containing evidence; 0.8–0.7 for highly useful content; 0.6–0.3 for weakly related background; 0.2-0.0 for irrelevant text.}. A chunk (c$_i$) is counted as relevant if its score exceeds this threshold\footnote{The threshold is set to $\tau=0.7$ to include only evidence judged highly useful.}, and the final score is computed as:
\begin{equation}
    \text{RetrievalRelevance} =
    \frac{\#\{\, c_i \ge \tau \,\}}{N}
\end{equation}
where $N$ is the number of retrieved chunks. This precision-style formulation is simple, cost-efficient, and provides actionable diagnostic insight into retrieval quality. Importantly, it evaluates the usefulness of retrieved documents directly from the question alone, without requiring reference answers. 

\subsubsection{Retrieval Coverage}\label{RC}
This metric measures whether the retrieved contexts (C) collectively contain the evidence needed to answer the question (Q), without requiring reference contexts. We first derive a small set of atomic aspects from the question alone and reuse the same aspects across metrics for consistency. Let $\mathcal{A}$ denote this set of aspects and let $\mathcal{R}_{\text{cov}} \subseteq \mathcal{A}$ be the subset of aspects supported by at least one retrieved document. The corresponding score is:
\begin{equation}
    \text{RetrievalCoverage} =
    \frac{|\mathcal{R}_{\text{cov}}|}{|\mathcal{A}|}
\end{equation}
This recall-style metric indicates whether the retriever has surfaced enough evidence to cover all parts of the question.  

\subsubsection{Clarity}\label{C}
This metric evaluates the linguistic quality of the generated answer (A), assessing grammar, fluency, logical flow, conciseness, and overall readability. A single LLM call returns a score in $[0,1]$ along with a brief explanation and suggested improvements. Short answers are also checked for naturalness and readability. Overall, this metric provides a compact indication of how clearly the answer is written.

\subsubsection{Answer Relevance}\label{AR}
Answer Relevance measures how well the generated answer (A) aligns with the user's question (Q) intent. The metric considers only the question and the generated answer, and assigns a score in $[0,1]$ based on topical focus and whether the answer meaningfully addresses what the question is asking. It ignores factual correctness and stylistic quality. High scores thus indicate that the answer stays on-topic and captures the main intent, while lower scores reflect partial, generic, or off-topic content. The judge additionally returns short lists of missing or off-topic parts to provide interpretable signals about alignment.

\subsubsection{Answer Completeness}\label{AC}
Answer Completeness measures how well the answer covers the different aspects implied by the question. Using the same aspect set $\mathcal{A}$ (from~\ref{RC}), the metric checks each aspect against the answer and identifies the subset $\mathcal{A}_{\text{cov}} \subseteq \mathcal{A}$, i.e., the aspects that the answer explicitly addresses (optionally with short supporting snippets). The final score is computed as:
\begin{equation}
    \text{AnswerCompleteness} =
    \frac{|\mathcal{A}_{\text{cov}}|}{|\mathcal{A}|}
\end{equation}
This reference-free metric captures how thoroughly the answer resolves the information needs expressed by the question. 

\subsubsection{Strict Faithfulness}\label{SF}
To assess factual grounding, we introduce a \textit{single-pass faithfulness metric} that determines whether the retrieved context explicitly supports each claim in a generated answer. The metric uses an LLM-generated structured analysis that jointly (1) decomposes the answer into minimal factual claims, and (2) classifies each claim as \textit{supported}, \textit{partially hallucinated}, or \textit{fully hallucinated}. Unlike multi-stage evaluation pipelines that require multiple LLM calls or rely on heuristic post-processing, our approach embeds strict verification rules directly within a single prompt. These rules require exact surface-form matches for \textit{key entities} (e.g., people, locations, organizations) and exact agreement for \textit{temporal expressions} such as years and dates. Claims containing unsupported or contradictory information are labeled as hallucinated. The final faithfulness score is computed as:
\begin{equation}
    \text{ StrictFaithfulness} = \frac{|\mathcal{C}_{\text{supported}}|}{|\mathcal{C}_{\text{supported}}| + |\mathcal{C}_{\text{hallucinated}}|}
\end{equation}
where $\mathcal{C}_{\text{supported}}$ is the set of claims fully grounded in the context, and $\mathcal{C}_{\text{hallucinated}}$ is the set of claims marked as partially or fully hallucinated. The result is a low-cost, transparent score that helps users quickly see which parts of the answer are actually backed by evidence. 

\subsubsection{Generic Calibration} \label{Cali.}
LLM-based evaluators are sensitive to sampling noise, decoding temperature, and model choice, yet most RAG evaluation pipelines implicitly assume that a single judge is both stable and trustworthy. To make this assumption explicit, we introduce a \emph{generic calibration metric} that quantifies agreement across multiple judge configurations and can be applied to any \textsc{RAGVue} metric. For each evaluation case, we run the same underlying metric under several $(\text{model},\,\text{temperature})$ configurations and obtain a set of scores $s_1,\ldots,s_k$. We define calibration agreement as
\begin{equation}
    \text{Calibration} = 1 - (\max_i s_i - \min_i s_i),
\end{equation}
which assigns high values when judges behave consistently and low values when their outputs diverge. This formulation captures cross-model uncertainty and makes explicit when a metric's output is robust versus brittle. The calibration metric not only reports an aggregate agreement score but also surfaces per-judge explanations, allowing users to trace which configurations disagree, by how much, and on what basis. As a result, \textsc{RAGVue} exposes the reliability of evaluation signals themselves, which helps to detect unstable judgments, model-induced variance, or temperature sensitivity in any metric used within the evaluation pipeline.

\subsection{Operational Modes \& Availability}
Our evaluation framework supports two complementary operational modes and user interfaces, enabling flexible integration into research workflows and production pipelines.

\subsubsection{Operational Modes}
\textsc{RAGVue} provides two modes for users. In \textbf{manual mode}, users control which metrics are executed and how results are aggregated, offering full transparency and fine-grained control. In \textbf{agentic mode}, an internal orchestration agent fully automates evaluation. The agent selects appropriate retrieval and answer-level metrics based on the presence of context, the availability of an answer, and the user query. It then executes these metrics in a single pass and synthesizes high-level scores, including an overall retrieval score (harmonic mean of relevance and coverage) and an answer-level composite score (weighted blend of strict faithfulness, relevance, completeness, and clarity).

\subsubsection{Availability}
\textsc{RAGVue} is released under a Creative Commons (CC) license\footnote{Full details are available in our GitHub repository}, and can be used through multiple access modes depending on the user's preference and technical requirements.

\paragraph{Python API.}
\textsc{RAGVue} can also be used directly as a Python library (Fig.~\ref{fig:RAGVue-python-api}). Users import the evaluator, load a JSONL dataset, and run all metrics with a single function call, making this mode ideal for integration into notebooks, scripts, and automated pipelines.


\begin{figure}[ht]
\centering
\begin{lstlisting}[style=ragvue, language=Python]
from ragvue import evaluate, load_metrics
items = [
    {"question": "...", "answer": "...",
    "context": [...]}]
metrics = load_metrics().keys()
report = evaluate(items, metrics=list(metrics))
print(report)
\end{lstlisting}
\caption{\textsc{RAGVue} Python API usage example.}
\label{fig:RAGVue-python-api}
\end{figure}

\paragraph{Python/Command-Line Interface (CLI).}
\textsc{RAGVue} provides a simple command-line interface (ragvue-cli) for terminal-based workflows, as shown in Figure~\ref{fig:RAGVue-cli}. A lightweight Python runner (ragvue-py) is also available, as illustrated in Figure~\ref{fig:RAGVue-py}. Both interfaces support listing available metrics, running manual evaluations, and executing the agentic mode, enabling fast and scriptable evaluation without writing additional code.



\begin{figure}[ht]
\centering
\begin{lstlisting}[style=ragvue, language=bash]
# Help
ragvue-cli --help

# List all available metrics
ragvue-cli list-metrics

# Manual evaluation (choose metrics explicitly)
ragvue-cli eval \
  --inputs <your_data.jsonl> \
  --metrics <metrics> \
  --out-base report_manual \
  --format "json,md,csv"

# Agentic evaluation (auto-select metrics)
ragvue-cli agentic \
  --inputs <your_data.jsonl> \
  --out-base report_agentic \
  --format "json,md,csv"
\end{lstlisting}
\caption{Example usage of the \textsc{RAGVue} command-line interface (\texttt{ragvue-cli}).}
\label{fig:RAGVue-cli}
\end{figure}

\begin{figure}[ht]
\centering
\begin{lstlisting}[style=ragvue, language=bash]
# Help
ragvue-py --help

# Manual Mode Usage
ragvue-py --input <your_data> --metrics <metrics> \
  --out-base report_manual --skip-agentic

# Agentic Mode Usage
ragvue-py --input <your_data> --metrics <metrics> \
  --agentic-out report_agentic --skip-manual
\end{lstlisting}
\caption{Example usage of the \textsc{RAGVue} chat Python command-line runner (\texttt{ragvue-py}).}
\label{fig:RAGVue-py}
\end{figure}

\paragraph{Local Streamlit Application.}
For no-code, interactive usage, we provide a Streamlit-based UI that exposes the same capabilities through an interactive browser interface. The application is run locally: users clone the repository and start the interface with a standard command such as \texttt{streamlit run streamlit\_app.py}. Within this local UI, users can upload JSONL files, select operational modes, set/paste API keys for the current session, and generate formatted reports without writing code. This interface is targeted at practitioners who prefer a point-and-click workflow while keeping all data and keys on their own machine. The images of our UI are shown in Appendix~\ref{appendix_sec:UI}

\section{Experiments \& Discussion}

\subsection{Dataset}\label{sec:synthetic}
We construct our evaluation dataset based on the multihop StrategyQA~\cite{geva-etal-2021-aristotle} benchmark. Each item contains a question, the reference yes/no label, the supporting facts, the decomposition steps, and the Wikipedia evidence titles. The supporting facts are cleaned and used as independent context snippets. In the next stage, we generate five answer variants for each question, such as ideal, partial, unclear, off-topic, and hallucinated. These variants capture common RAG failure modes by altering the correctness of the label, the completeness of the explanation, and the relevance or confidence of the response. Each answer is stored with its associated metadata (question ID, reference label, contexts, supporting facts, decomposition, and evidence titles), producing exactly five synthetic examples per question. For this study, we created 100 synthetic $(Q, C, A)$ triplets from StrategyQA. The final dataset is exported in two formats: a RAGAS-compatible JSON for metric-based evaluation and a \textsc{RAGVue}-ready JSONL for interactive inspection.

\subsection{\textsc{RAGVue} vs.\ RAGAS Performance }

\paragraph{Computational Time.} We first measured latency on the 100 queries described in Section~\ref{sec:synthetic}. RAGAS averaged 18.26 seconds per query, while \textsc{RAGVue} averaged 18.87 seconds. This represents a marginal 3.4\% increase in per-item latency, which is negligible. As shown in the boxplot (Appendix~\ref{app:time}), both systems exhibit nearly identical latency distributions, including similar medians, inter-quartile ranges, and outliers. Importantly, \textsc{RAGVue} provides richer diagnostics, and this enhanced granularity makes it more actionable for system debugging and improvements, rendering the small computational overhead a worthwhile trade-off.

\paragraph{Quantitative Analysis.}
We next summarize the behavior of both evaluators using descriptive statistics over the 100 queries (Appendix~\ref{app:quantitative}). RAGAS faithfulness has a mean of 0.52, while answer relevance and response groundedness average at 0.24 and 0.39, respectively, indicating that RAGAS often judges answers as only weakly relevant or weakly grounded. In contrast, \textsc{RAGVue} reports low average answer completeness (0.12) and moderate answer relevance (0.37), alongside consistently high clarity scores (0.70). Its retrieval metrics center around 0.50 for coverage and 0.42 for relevance, while strict-faithfulness has a mean of 0.40, reflecting a wide range of partially supported and unsupported answers.

A correlation analysis (see Appendix~\ref{app:quantitative}) shows that the two evaluators align on broad, generation-focused behavior but diverge sharply on retrieval-focused metrics. RAGAS faithfulness, answer relevancy, and response groundedness correlate strongly with \textsc{RAGVue}'s strict faithfulness and answer relevance, indicating comparable sensitivity to high-level answer correctness. However, RAGAS's retrieval-related metrics, context relevance and response groundedness, correlate only weakly or inconsistently with \textsc{RAGVue}'s retrieval coverage and retrieval relevance. This reveals that RAGAS often conflates insufficient retrieval with unsupported reasoning, whereas \textsc{RAGVue} explicitly separates retrieval performance from generation performance.

\paragraph{Qualitative Analysis.}
Our qualitative inspection reveals several systematic failures that RAGAS does not diagnose, and some examples are presented in Appendix~\ref{appendix:qualitative}. We find that RAGAS often gives scores that look reasonable but do not explain why an answer fails. When the model gives vague, unsupported, or partially relevant answers, RAGAS may still assign high or mid-range faithfulness scores because it only checks for direct contradictions and does not account for missing multi-hop reasoning, unanswered parts of the question, or unsupported conclusions. \textsc{RAGVue}, on the other hand, clearly shows what went wrong: it marks claims as unsupported when the evidence does not back them, highlights when the answer ignores key aspects of the question, and indicates whether retrieval fully or only partially matched what was needed. As a result, \textsc{RAGVue} makes it easy to see whether the error comes from retrieval, grounding, or the model's reasoning, which is not possible with RAGAS. Overall, the qualitative analysis shows that \textsc{RAGVue} provides clearer and more actionable feedback for diagnosing RAG system failures.

\paragraph{Discussion.}Our quantitative and qualitative analyses directly reflect the structural limitations of RAGAS. Context Relevance collapses all chunks into a single aggregated score and cannot show which passages are missing or irrelevant. \textsc{RAGVue} provides per-chunk relevance without requiring reference answers, clearly exposing retrieval strengths and failures. Likewise, RAGAS's Context Recall requires a reference answer and multi-step alignment, while \textsc{RAGVue}'s Retrieval Coverage operates directly on the question and retrieved documents, making it usable even when references are unavailable.
On the generation side, RAGAS's Answer Relevancy relies on embedding-based synthetic question generation, capturing only coarse semantic overlap. \textsc{RAGVue}'s Answer Relevance is intent-aware and identifies missing or off-topic elements, yielding actionable diagnostics about why an answer may be incomplete. Finally, RAGAS's Response Groundedness assigns coarse labels (0/1/2) based on inferred support in the context, but cannot reveal which question aspects were addressed or missed. \textsc{RAGVue}'s Answer Completeness evaluates coverage directly from the question’s aspect structure, producing a fine-grained completeness signal. Strict Faithfulness exhibits the clearest contrast: RAGAS uses a two-step pipeline to extract statements from the answer and verify each with a semantic-inference prompt that tolerates semantic drift, whereas \textsc{RAGVue} decomposes the answer into atomic claims and enforces exact evidence matching for key entities and temporal expressions. This yields a stricter, more deterministic assessment of factual support. Overall, these results show that, while RAGAS provides high-level semantic judgments, \textsc{RAGVue} delivers finer-grained, retrieval-aware, and more diagnostically meaningful evaluation signals.

\section{Conclusion}
\textsc{RAGVue} provides an automated, diagnostic, and fully reference-free evaluation framework tailored for explainable assessment of RAG systems. It separates retrieval and generation-level metrics, delivers structured explanations rather than opaque scalar scores, and exposes the underlying causes of model failures. The agentic evaluation mode makes the framework immediately usable with minimal setup, automatically selecting appropriate metrics and producing structured reports that highlight where and why a pipeline breaks. By combining fine-grained metrics with transparent reasoning traces and cross-model calibration for reliability, \textsc{RAGVue} reveals whether an error stems from retrieval drift, missing evidence, unsupported reasoning, or incomplete answers, problems that traditional metrics such as RAGAS often conflate. Overall, \textsc{RAGVue} functions not only as an evaluator but as a practical debugging tool for real-world RAG development, that helps users identify weaknesses and iteratively improve their RAG systems.

\bibliography{references}

@article{geva-etal-2021-aristotle,
    title = "Did Aristotle Use a Laptop? A Question Answering Benchmark with Implicit Reasoning Strategies",
    author = "Geva, Mor  and
      Khashabi, Daniel  and
      Segal, Elad  and
      Khot, Tushar  and
      Roth, Dan  and
      Berant, Jonathan",
    editor = "Roark, Brian  and
      Nenkova, Ani",
    journal = "Transactions of the Association for Computational Linguistics",
    volume = "9",
    year = "2021",
    address = "Cambridge, MA",
    publisher = "MIT Press",
    url = "https://aclanthology.org/2021.tacl-1.21/",
    doi = "10.1162/tacl_a_00370",
    pages = "346--361"
}

@article{lewis2020retrieval,
  title={Retrieval-augmented generation for knowledge-intensive nlp tasks},
  author={Lewis, Patrick and Perez, Ethan and Piktus, Aleksandra and Petroni, Fabio and Karpukhin, Vladimir and Goyal, Naman and K{\"u}ttler, Heinrich and Lewis, Mike and Yih, Wen-tau and Rockt{\"a}schel, Tim and others},
  journal={Advances in neural information processing systems},
  volume={33},
  pages={9459--9474},
  year={2020}
}

@inproceedings{yu2024evaluation,
  title={Evaluation of retrieval-augmented generation: A survey},
  author={Yu, Hao and Gan, Aoran and Zhang, Kai and Tong, Shiwei and Liu, Qi and Liu, Zhaofeng},
  booktitle={CCF Conference on Big Data},
  pages={102--120},
  year={2024},
  organization={Springer}
}

@article{gan2025retrieval,
  title={Retrieval Augmented Generation Evaluation in the Era of Large Language Models: A Comprehensive Survey},
  author={Gan, Aoran and Yu, Hao and Zhang, Kai and Liu, Qi and Yan, Wenyu and Huang, Zhenya and Tong, Shiwei and Hu, Guoping},
  journal={arXiv preprint arXiv:2504.14891},
  year={2025}
}

@inproceedings{liu-etal-2023-g,
    title = "{G}-Eval: {NLG} Evaluation using Gpt-4 with Better Human Alignment",
    author = "Liu, Yang  and
      Iter, Dan  and
      Xu, Yichong  and
      Wang, Shuohang  and
      Xu, Ruochen  and
      Zhu, Chenguang",
    editor = "Bouamor, Houda  and
      Pino, Juan  and
      Bali, Kalika",
    booktitle = "Proceedings of the 2023 Conference on Empirical Methods in Natural Language Processing",
    month = dec,
    year = "2023",
    address = "Singapore",
    publisher = "Association for Computational Linguistics",
    url = "https://aclanthology.org/2023.emnlp-main.153/",
    doi = "10.18653/v1/2023.emnlp-main.153",
    pages = "2511--2522"
}

@inproceedings{es2024ragas,
  title={Ragas: Automated evaluation of retrieval augmented generation},
  author={Es, Shahul and James, Jithin and Anke, Luis Espinosa and Schockaert, Steven},
  booktitle={Proceedings of the 18th Conference of the European Chapter of the Association for Computational Linguistics: System Demonstrations},
  pages={150--158},
  year={2024}
}

@inproceedings{saad-falcon-etal-2024-ares,
    title = "{ARES}: An Automated Evaluation Framework for Retrieval-Augmented Generation Systems",
    author = "Saad-Falcon, Jon  and
      Khattab, Omar  and
      Potts, Christopher  and
      Zaharia, Matei",
    editor = "Duh, Kevin  and
      Gomez, Helena  and
      Bethard, Steven",
    booktitle = "Proceedings of the 2024 Conference of the North American Chapter of the Association for Computational Linguistics: Human Language Technologies (Volume 1: Long Papers)",
    month = jun,
    year = "2024",
    address = "Mexico City, Mexico",
    publisher = "Association for Computational Linguistics",
    url = "https://aclanthology.org/2024.naacl-long.20/",
    doi = "10.18653/v1/2024.naacl-long.20",
    pages = "338--354"
}

@article{ru2024ragchecker,
  title={Ragchecker: A fine-grained framework for diagnosing retrieval-augmented generation},
  author={Ru, Dongyu and Qiu, Lin and Hu, Xiangkun and Zhang, Tianhang and Shi, Peng and Chang, Shuaichen and Jiayang, Cheng and Wang, Cunxiang and Sun, Shichao and Li, Huanyu and others},
  journal={Advances in Neural Information Processing Systems},
  volume={37},
  pages={21999--22027},
  year={2024}
}

@inproceedings{niu2024ragtruth,
  title={Ragtruth: A hallucination corpus for developing trustworthy retrieval-augmented language models},
  author={Niu, Cheng and Wu, Yuanhao and Zhu, Juno and Xu, Siliang and Shum, Kashun and Zhong, Randy and Song, Juntong and Zhang, Tong},
  booktitle={Proceedings of the 62nd Annual Meeting of the Association for Computational Linguistics (Volume 1: Long Papers)},
  pages={10862--10878},
  year={2024}
}

@article{schroeder2024can,
  title={Can you trust llm judgments? reliability of llm-as-a-judge},
  author={Schroeder, Kayla and Wood-Doughty, Zach},
  journal={arXiv preprint arXiv:2412.12509},
  year={2024}
}

@inproceedings{cruz-blandon-etal-2025-memerag,
    title = "{MEMERAG}: A Multilingual End-to-End Meta-Evaluation Benchmark for Retrieval Augmented Generation",
    author = "Cruz Bland{\'o}n, Mar{\'i}a Andrea  and
      Talur, Jayasimha  and
      Charron, Bruno  and
      Liu, Dong  and
      Mansour, Saab  and
      Federico, Marcello",
    editor = "Che, Wanxiang  and
      Nabende, Joyce  and
      Shutova, Ekaterina  and
      Pilehvar, Mohammad Taher",
    booktitle = "Proceedings of the 63rd Annual Meeting of the Association for Computational Linguistics (Volume 1: Long Papers)",
    month = jul,
    year = "2025",
    address = "Vienna, Austria",
    publisher = "Association for Computational Linguistics",
    url = "https://aclanthology.org/2025.acl-long.1101/",
    doi = "10.18653/v1/2025.acl-long.1101",
    pages = "22577--22595",
    ISBN = "979-8-89176-251-0"
}

@inproceedings{manakul-etal-2023-selfcheckgpt,
    title = "{S}elf{C}heck{GPT}: Zero-Resource Black-Box Hallucination Detection for Generative Large Language Models",
    author = "Manakul, Potsawee  and
      Liusie, Adian  and
      Gales, Mark",
    editor = "Bouamor, Houda  and
      Pino, Juan  and
      Bali, Kalika",
    booktitle = "Proceedings of the 2023 Conference on Empirical Methods in Natural Language Processing",
    month = dec,
    year = "2023",
    address = "Singapore",
    publisher = "Association for Computational Linguistics",
    url = "https://aclanthology.org/2023.emnlp-main.557/",
    doi = "10.18653/v1/2023.emnlp-main.557",
    pages = "9004--9017"
}

@article{song2024measuring,
  title={Measuring and enhancing trustworthiness of LLMs in RAG through grounded attributions and learning to refuse},
  author={Song, Maojia and Sim, Shang Hong and Bhardwaj, Rishabh and Chieu, Hai Leong and Majumder, Navonil and Poria, Soujanya},
  journal={arXiv preprint arXiv:2409.11242},
  year={2024}
}

@inproceedings{zhu-etal-2025-rageval,
    title = "{RAGE}val: Scenario Specific {RAG} Evaluation Dataset Generation Framework",
    author = "Zhu, Kunlun  and
      Luo, Yifan  and
      Xu, Dingling  and
      Yan, Yukun  and
      Liu, Zhenghao  and
      Yu, Shi  and
      Wang, Ruobing  and
      Wang, Shuo  and
      Li, Yishan  and
      Zhang, Nan  and
      Han, Xu  and
      Liu, Zhiyuan  and
      Sun, Maosong",
    editor = "Che, Wanxiang  and
      Nabende, Joyce  and
      Shutova, Ekaterina  and
      Pilehvar, Mohammad Taher",
    booktitle = "Proceedings of the 63rd Annual Meeting of the Association for Computational Linguistics (Volume 1: Long Papers)",
    month = jul,
    year = "2025",
    address = "Vienna, Austria",
    publisher = "Association for Computational Linguistics",
    url = "https://aclanthology.org/2025.acl-long.418/",
    doi = "10.18653/v1/2025.acl-long.418",
    pages = "8520--8544",
    ISBN = "979-8-89176-251-0"
}

@inproceedings{zeng-etal-2025-rag,
    title = "{S}-{RAG}: A Novel Audit Framework for Detecting Unauthorized Use of Personal Data in {RAG} Systems",
    author = "Zeng, Zhirui  and
      Liu, Jiamou  and
      Chiang, Meng-Fen  and
      He, Jialing  and
      Zhang, Zijian",
    editor = "Che, Wanxiang  and
      Nabende, Joyce  and
      Shutova, Ekaterina  and
      Pilehvar, Mohammad Taher",
    booktitle = "Proceedings of the 63rd Annual Meeting of the Association for Computational Linguistics (Volume 1: Long Papers)",
    month = jul,
    year = "2025",
    address = "Vienna, Austria",
    publisher = "Association for Computational Linguistics",
    url = "https://aclanthology.org/2025.acl-long.512/",
    doi = "10.18653/v1/2025.acl-long.512",
    pages = "10375--10385",
    ISBN = "979-8-89176-251-0"
}

@inproceedings{li-etal-2025-rag,
    title = "{RAG}-Zeval: Enhancing {RAG} Responses Evaluator through End-to-End Reasoning and Ranking-Based Reinforcement Learning",
    author = "Li, Kun  and
      Li, Yunxiang  and
      Zhang, Tianhua  and
      Luo, Hongyin  and
      Wu, Xixin  and
      Glass, James R.  and
      Meng, Helen M.",
    editor = "Christodoulopoulos, Christos  and
      Chakraborty, Tanmoy  and
      Rose, Carolyn  and
      Peng, Violet",
    booktitle = "Proceedings of the 2025 Conference on Empirical Methods in Natural Language Processing",
    month = nov,
    year = "2025",
    address = "Suzhou, China",
    publisher = "Association for Computational Linguistics",
    url = "https://aclanthology.org/2025.emnlp-main.1267/",
    doi = "10.18653/v1/2025.emnlp-main.1267",
    pages = "24936--24954",
    ISBN = "979-8-89176-332-6"
}

@inproceedings{peng-etal-2025-unanswerability,
    title = "Unanswerability Evaluation for Retrieval Augmented Generation",
    author = "Peng, Xiangyu  and
      Choubey, Prafulla Kumar  and
      Xiong, Caiming  and
      Wu, Chien-Sheng",
    editor = "Che, Wanxiang  and
      Nabende, Joyce  and
      Shutova, Ekaterina  and
      Pilehvar, Mohammad Taher",
    booktitle = "Proceedings of the 63rd Annual Meeting of the Association for Computational Linguistics (Volume 1: Long Papers)",
    month = jul,
    year = "2025",
    address = "Vienna, Austria",
    publisher = "Association for Computational Linguistics",
    url = "https://aclanthology.org/2025.acl-long.415/",
    doi = "10.18653/v1/2025.acl-long.415",
    pages = "8452--8472",
    ISBN = "979-8-89176-251-0"
}

@article{10.1162/TACL.a.19,
    author = {Katsis, Yannis and Rosenthal, Sara and Fadnis, Kshitij and Gunasekara, Chulaka and Lee, Young-Suk and Popa, Lucian and Shah, Vraj and Zhu, Huaiyu and Contractor, Danish and Danilevsky, Marina},
    title = {mtRAG: A Multi-Turn Conversational Benchmark for Evaluating
                    Retrieval-Augmented Generation Systems},
    journal = {Transactions of the Association for Computational Linguistics},
    volume = {13},
    pages = {784-808},
    year = {2025},
    month = {07},
    issn = {2307-387X},
    doi = {10.1162/TACL.a.19},
    url = {https://doi.org/10.1162/TACL.a.19},
    eprint = {https://direct.mit.edu/tacl/article-pdf/doi/10.1162/TACL.a.19/2540217/tacl.a.19.pdf},
}

@inproceedings{ouyang-etal-2025-hoh,
    title = "{H}o{H}: A Dynamic Benchmark for Evaluating the Impact of Outdated Information on Retrieval-Augmented Generation",
    author = "Ouyang, Jie  and
      Pan, Tingyue  and
      Cheng, Mingyue  and
      Yan, Ruiran  and
      Luo, Yucong  and
      Lin, Jiaying  and
      Liu, Qi",
    editor = "Che, Wanxiang  and
      Nabende, Joyce  and
      Shutova, Ekaterina  and
      Pilehvar, Mohammad Taher",
    booktitle = "Proceedings of the 63rd Annual Meeting of the Association for Computational Linguistics (Volume 1: Long Papers)",
    month = jul,
    year = "2025",
    address = "Vienna, Austria",
    publisher = "Association for Computational Linguistics",
    url = "https://aclanthology.org/2025.acl-long.301/",
    doi = "10.18653/v1/2025.acl-long.301",
    pages = "6036--6063",
    ISBN = "979-8-89176-251-0"
}

@inproceedings{sorodoc-etal-2025-garage,
    title = "{G}a{RAG}e: A Benchmark with Grounding Annotations for {RAG} Evaluation",
    author = "Sorodoc, Ionut Teodor  and
      Ribeiro, Leonardo F. R.  and
      Blloshmi, Rexhina  and
      Davis, Christopher  and
      de Gispert, Adri{\`a}",
    editor = "Che, Wanxiang  and
      Nabende, Joyce  and
      Shutova, Ekaterina  and
      Pilehvar, Mohammad Taher",
    booktitle = "Findings of the Association for Computational Linguistics: ACL 2025",
    month = jul,
    year = "2025",
    address = "Vienna, Austria",
    publisher = "Association for Computational Linguistics",
    url = "https://aclanthology.org/2025.findings-acl.875/",
    doi = "10.18653/v1/2025.findings-acl.875",
    pages = "17030--17049",
    ISBN = "979-8-89176-256-5"
}

@article{rosenthal-etal-2025-clapnq,
    title = "{CLAP}nq: Cohesive Long-form Answers from Passages in Natural Questions for {RAG} systems",
    author = "Rosenthal, Sara  and
      Sil, Avirup  and
      Florian, Radu  and
      Roukos, Salim",
    journal = "Transactions of the Association for Computational Linguistics",
    volume = "13",
    year = "2025",
    address = "Cambridge, MA",
    publisher = "MIT Press",
    url = "https://aclanthology.org/2025.tacl-1.3/",
    doi = "10.1162/tacl_a_00729",
    pages = "53--72"
}

@inproceedings{guu2020retrieval,
  title={Retrieval augmented language model pre-training},
  author={Guu, Kelvin and Lee, Kenton and Tung, Zora and Pasupat, Panupong and Chang, Mingwei},
  booktitle={International conference on machine learning},
  pages={3929--3938},
  year={2020},
  organization={PMLR}
}

@inproceedings{kocmi-federmann-2023-large,
    title = "Large Language Models Are State-of-the-Art Evaluators of Translation Quality",
    author = "Kocmi, Tom  and
      Federmann, Christian",
    editor = "Nurminen, Mary  and
      Brenner, Judith  and
      Koponen, Maarit  and
      Latomaa, Sirkku  and
      Mikhailov, Mikhail  and
      Schierl, Frederike  and
      Ranasinghe, Tharindu  and
      Vanmassenhove, Eva  and
      Vidal, Sergi Alvarez  and
      Aranberri, Nora  and
      Nunziatini, Mara  and
      Escart{\'i}n, Carla Parra  and
      Forcada, Mikel  and
      Popovic, Maja  and
      Scarton, Carolina  and
      Moniz, Helena",
    booktitle = "Proceedings of the 24th Annual Conference of the European Association for Machine Translation",
    month = jun,
    year = "2023",
    address = "Tampere, Finland",
    publisher = "European Association for Machine Translation",
    url = "https://aclanthology.org/2023.eamt-1.19/",
    pages = "193--203"
}

@inproceedings{wang-etal-2023-chatgpt,
    title = "Is {C}hat{GPT} a Good {NLG} Evaluator? A Preliminary Study",
    author = "Wang, Jiaan  and
      Liang, Yunlong  and
      Meng, Fandong  and
      Sun, Zengkui  and
      Shi, Haoxiang  and
      Li, Zhixu  and
      Xu, Jinan  and
      Qu, Jianfeng  and
      Zhou, Jie",
    editor = "Dong, Yue  and
      Xiao, Wen  and
      Wang, Lu  and
      Liu, Fei  and
      Carenini, Giuseppe",
    booktitle = "Proceedings of the 4th New Frontiers in Summarization Workshop",
    month = dec,
    year = "2023",
    address = "Singapore",
    publisher = "Association for Computational Linguistics",
    url = "https://aclanthology.org/2023.newsum-1.1/",
    doi = "10.18653/v1/2023.newsum-1.1",
    pages = "1--11"
}

@article{khattab2021relevance,
  title={Relevance-guided supervision for openqa with colbert},
  author={Khattab, Omar and Potts, Christopher and Zaharia, Matei},
  journal={Transactions of the association for computational linguistics},
  volume={9},
  pages={929--944},
  year={2021},
  publisher={MIT Press One Rogers Street, Cambridge, MA 02142-1209, USA journals-info~…}
}

@article{10.5555/3648699.3648950,
author = {Izacard, Gautier and Lewis, Patrick and Lomeli, Maria and Hosseini, Lucas and Petroni, Fabio and Schick, Timo and Dwivedi-Yu, Jane and Joulin, Armand and Riedel, Sebastian and Grave, Edouard},
title = {Atlas: few-shot learning with retrieval augmented language models},
year = {2023},
issue_date = {January 2023},
publisher = {JMLR.org},
volume = {24},
number = {1},
issn = {1532-4435},
journal = {J. Mach. Learn. Res.},
month = jan,
articleno = {251},
numpages = {43}
}

@inproceedings{10.5555/3666122.3668142,
author = {Zheng, Lianmin and Chiang, Wei-Lin and Sheng, Ying and Zhuang, Siyuan and Wu, Zhanghao and Zhuang, Yonghao and Lin, Zi and Li, Zhuohan and Li, Dacheng and Xing, Eric P. and Zhang, Hao and Gonzalez, Joseph E. and Stoica, Ion},
title = {Judging LLM-as-a-judge with MT-bench and Chatbot Arena},
year = {2023},
publisher = {Curran Associates Inc.},
address = {Red Hook, NY, USA},
booktitle = {Proceedings of the 37th International Conference on Neural Information Processing Systems},
articleno = {2020},
numpages = {29},
location = {New Orleans, LA, USA},
series = {NIPS '23}
}

@inproceedings{liu-etal-2024-calibrating,
    title = "Calibrating {LLM}-Based Evaluator",
    author = "Liu, Yuxuan  and
      Yang, Tianchi  and
      Huang, Shaohan  and
      Zhang, Zihan  and
      Huang, Haizhen  and
      Wei, Furu  and
      Deng, Weiwei  and
      Sun, Feng  and
      Zhang, Qi",
    editor = "Calzolari, Nicoletta  and
      Kan, Min-Yen  and
      Hoste, Veronique  and
      Lenci, Alessandro  and
      Sakti, Sakriani  and
      Xue, Nianwen",
    booktitle = "Proceedings of the 2024 Joint International Conference on Computational Linguistics, Language Resources and Evaluation (LREC-COLING 2024)",
    month = may,
    year = "2024",
    address = "Torino, Italia",
    publisher = "ELRA and ICCL",
    url = "https://aclanthology.org/2024.lrec-main.237/",
    pages = "2638--2656"
}

@inproceedings{liu-etal-2025-judge,
    title = "Judge as A Judge: Improving the Evaluation of Retrieval-Augmented Generation through the Judge-Consistency of Large Language Models",
    author = "Liu, Shuliang  and
      Li, Xinze  and
      Liu, Zhenghao  and
      Yan, Yukun  and
      Yang, Cheng  and
      Zeng, Zheni  and
      Liu, Zhiyuan  and
      Sun, Maosong  and
      Yu, Ge",
    editor = "Che, Wanxiang  and
      Nabende, Joyce  and
      Shutova, Ekaterina  and
      Pilehvar, Mohammad Taher",
    booktitle = "Findings of the Association for Computational Linguistics: ACL 2025",
    month = jul,
    year = "2025",
    address = "Vienna, Austria",
    publisher = "Association for Computational Linguistics",
    url = "https://aclanthology.org/2025.findings-acl.301/",
    doi = "10.18653/v1/2025.findings-acl.301",
    pages = "5788--5807",
    ISBN = "979-8-89176-256-5"
}

@inproceedings{10.5555/3737916.3740113,
author = {Panickssery, Arjun and Bowman, Samuel R. and Feng, Shi},
title = {LLM evaluators recognize and favor their own generations},
year = {2024},
isbn = {9798331314385},
publisher = {Curran Associates Inc.},
address = {Red Hook, NY, USA},
booktitle = {Proceedings of the 38th International Conference on Neural Information Processing Systems},
articleno = {2197},
numpages = {31},
location = {Vancouver, BC, Canada},
series = {NIPS '24}
}

@inproceedings{min-etal-2023-factscore,
    title = "{FA}ct{S}core: Fine-grained Atomic Evaluation of Factual Precision in Long Form Text Generation",
    author = "Min, Sewon  and
      Krishna, Kalpesh  and
      Lyu, Xinxi  and
      Lewis, Mike  and
      Yih, Wen-tau  and
      Koh, Pang  and
      Iyyer, Mohit  and
      Zettlemoyer, Luke  and
      Hajishirzi, Hannaneh",
    editor = "Bouamor, Houda  and
      Pino, Juan  and
      Bali, Kalika",
    booktitle = "Proceedings of the 2023 Conference on Empirical Methods in Natural Language Processing",
    month = dec,
    year = "2023",
    address = "Singapore",
    publisher = "Association for Computational Linguistics",
    url = "https://aclanthology.org/2023.emnlp-main.741/",
    doi = "10.18653/v1/2023.emnlp-main.741",
    pages = "12076--12100"
}

\appendix

\section{Summary of Metrics}\label{app:summary_metrics}
Our summary of metrics is available in Table~\ref{tab:RAGVue-metrics-summary}

\section{Computational Time plot} \label{app:time}
\begin{figure}[ht]
  \centering
  \includegraphics[width=\linewidth]{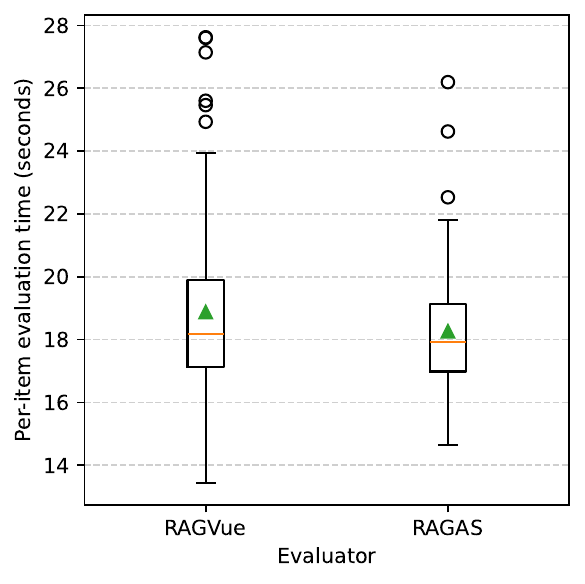}
  \caption{Distribution of per-item evaluation time for \textsc{RAGVue} and RAGAS on the 100-query benchmark.}
  \label{fig:eval_time_boxplot}
\end{figure}

The computational time box plot is shown in Figure~\ref{fig:eval_time_boxplot}

\section{Quantitative Analysis}\label{app:quantitative}
The descriptive statistics are shown in Table~\ref{tab:desc_ragas_RAGVue}, and the correlation results are provided in Table~\ref{tab:corr_ragas_RAGVue}.

\begin{table}[ht]
\centering
\small
\begin{tabular}{llrr}
\toprule
\textbf{System} & \textbf{Metric} & \textbf{Mean} & \textbf{Std} \\
\midrule
\multirow{4}{*}{RAGAS}
  & \texttt{faithfulness}           & 0.521 & 0.403 \\
  & \texttt{answer\_relevancy}      & 0.240 & 0.307 \\
  & \texttt{context\_relevance}     & 0.550 & 0.264 \\
  & \texttt{response\_groundedness} & 0.390 & 0.460 \\
\midrule
\multirow{6}{*}{\textsc{RAGVue}}
  & \texttt{answer\_completeness} & 0.121 & 0.225 \\
  & \texttt{answer\_relevance}    & 0.372 & 0.255 \\
  & \texttt{clarity}              & 0.698 & 0.100 \\
  & \texttt{retrieval\_coverage}  & 0.503 & 0.279 \\
  & \texttt{retrieval\_relevance} & 0.420 & 0.316 \\
  & \texttt{strict\_faithfulness} & 0.400 & 0.492 \\
\bottomrule
\end{tabular}
\caption{Descriptive statistics of RAGAS and \textsc{RAGVue} metrics on the 100-query benchmark (mean and standard deviation).}
\label{tab:desc_ragas_RAGVue}
\end{table}

\begin{table*}[ht]
\centering
\small
\caption{Spearman correlation between RAGAS metrics and \textsc{RAGVue} metrics on the 100-query benchmark.}
\label{tab:corr_ragas_RAGVue}
\begin{tabular}{lrrrrrr}
\toprule
 & \multicolumn{6}{c}{\textbf{\textsc{RAGVue} metrics}} \\
\cmidrule(lr){2-7}
\textbf{RAGAS} &
\texttt{ans\_comp.} &
\texttt{ans\_rel} &
\texttt{clarity} &
\texttt{ret\_cov} &
\texttt{ret\_rel} &
\texttt{strict\_faith} \\
\midrule
\texttt{faithfulness}  & 0.553 & 0.668 & 0.094 & 0.298 & 0.025 & 0.739 \\
\texttt{answer\_relevancy} & 0.704 & 0.644 & 0.172 & 0.193 & 0.053 & 0.958 \\
\texttt{context\_relevance} & 0.082 & 0.062 & 0.171 & 0.006 & 0.708 & 0.035 \\
\texttt{response\_groundedness} & 0.373 & 0.174 & 0.071 & 0.071 & 0.124 & 0.940 \\
\bottomrule
\end{tabular}
\end{table*}


\begin{table*}[t]
\centering
\small
\caption{Qualitative examples comparing RAGAS and \textsc{RAGVue} on real evaluation outputs.}
\label{tab:qual_examples}
\begin{tabular}{p{4.5cm}p{4.5cm}p{2.7cm}p{2.7cm}}
\toprule
\textbf{Question} &
\textbf{Model Answer} &
\textbf{RAGAS Scores} &
\textbf{\textsc{RAGVue} Scores} \\
\midrule

Are more people today related to Genghis Khan than Julius Caesar? &
``No, even though there is no strong supporting evidence…'' &
Faithfulness: 1.0 \newline
Answer rel.: 0.0 \newline
Context rel.: 0.5 \newline
Groundedness: 0.0 &
Strict faith.: 0.0 \newline
Answer rel.: 0.40 \newline
Completeness: 0.00 \newline
Ret. cov.: 0.33 \newline
Ret. rel.: 0.33 \\

\midrule
Could the members of \textit{The Police} perform lawful arrests? &
``Yes, even though there is no strong supporting evidence…'' &
Faithfulness: 0.5 \newline
Answer rel.: 0.0 \newline
Context rel.: 1.0 \newline
Groundedness: 0.0 &
Strict faith.: 0.0 \newline
Answer rel.: 0.40 \newline
Completeness: 0.00 \newline
Ret. cov.: 0.33 \newline
Ret. rel.: 1.00 \\

\midrule
Would a dog respond to a bell before a grey seal? &
``It is hard to say… probably true. The evidence is not entirely clear.'' &
Faithfulness: 0.67 \newline
Answer rel.: 0.0 \newline
Context rel.: 0.5 \newline
Groundedness: 0.0 &
Strict faith.: 0.0 \newline
Answer rel.: 0.40 \newline
Completeness: 0.00 \newline
Ret. cov.: 0.50 \newline
Ret. rel.: 0.33 \\
\bottomrule
\end{tabular}
\end{table*}

\section{Qualitative Case Studies - Examples} \label{appendix:qualitative}

Table~\ref{tab:qual_examples} provides item-level analyses for three representative evaluation cases drawn from our dataset. For each item, we report the behaviors of RAGAS and \textsc{RAGVue} alongside the structured diagnostic signals \textsc{RAGVue} generates.

\paragraph{Example 1.}
The model gives a vague answer (``No, even though there is no strong supporting evidence’’) that does not actually address the comparative question or follow from the retrieved facts. RAGAS still gives it a perfect faithfulness score (1.0) because it only checks for surface-level consistency and does not verify whether the conclusion is supported across multiple pieces of evidence, missing the needed \textbf{multi-hop reasoning}. \textsc{RAGVue} instead marks the claim as fully unsupported (strict faithfulness~$=0.0$), shows that none of the key parts of the question are answered (completeness~$=0.0$), and indicates that retrieval only partly matched what was needed (retrieval coverage~$=0.33$). Together, these signals clearly reveal an unsupported reasoning error that RAGAS fails to catch.

\paragraph{Example 2.}
The system retrieves the right information, i.e, both RAGAS and \textsc{RAGVue} show that the context is fully relevant. But the model still gives an incorrect answer (“Yes, even though there is no strong supporting evidence…”), which is not backed by the retrieved facts. RAGAS gives a mid-range faithfulness score (0.5) without explaining where the mistake comes from. \textsc{RAGVue} makes this clear: it marks the claim as completely unsupported (strict faithfulness~$=0.0$), shows that the answer covers none of the required points (completeness~$=0.0$), and confirms that retrieval was correct. This directly identifies the problem as a \emph{generation error}, not a retrieval issue.

\paragraph{Example 3.}
The model gives a vague answer (``probably true'') even though the retrieved evidence does not actually say which of the two (dog or grey seal) would respond first. RAGAS gives the answer a mid-range faithfulness score ($\approx 0.67$) because it does not directly contradict any single fact, but this does not explain what went wrong. \textsc{RAGVue} makes the issue clear: it marks the answer as fully unsupported (strict faithfulness~$=0.0$), shows that the model did not address the key parts of the question (completeness~$=0.0$), and indicates that only part of the retrieved information was relevant. As a result, \textsc{RAGVue} pinpoints that the failure also comes from the model’s reasoning, not only from retrieval, which is something RAGAS cannot show.

Across all three examples, \textsc{RAGVue} provides structured diagnostics that clearly distinguish whether errors stem from retrieval, grounding, or reasoning. In contrast, RAGAS offers only scalar scores, which obscure these distinctions in practice.


\renewcommand{\arraystretch}{1.0}
\begin{table*}
\centering
\small
\begin{tabular}{p{3.1cm} p{2cm} p{9.5cm}}
\toprule
\textbf{Metric} & \textbf{Inputs} & \textbf{What it Measures} \\
\midrule

\rowcolor{black!5}
\multicolumn{3}{c}{\textbf{\textcolor{blue}{Retrieval Metrics}}} \\
\addlinespace[0.4em]

\textit{Retrieval Relevance} & Q, C &
Evaluates how useful each retrieved chunk is for addressing the information needs of the question, based on per-chunk relevance scoring. \\

\textit{Retrieval Coverage} & Q, C &
Assesses whether the retrieved context collectively provides sufficient coverage for all sub-aspects required to answer the question. \\[0.25em]
\addlinespace[0.5em]

\rowcolor{black!5}
\multicolumn{3}{c}{\textbf{\textcolor{blue}{Answer  Metrics}}} \\
\addlinespace[0.4em]

\textit{Answer Relevance} & Q, A &
Measures how well the answer aligns with the intent and scope of the question, identifying missing, irrelevant, or off-topic content. \\

\textit{Answer Completeness} & Q, A &
Determines whether the answer fully addresses all aspects of the question without omissions. \\

\textit{Clarity} & A &
Evaluates the linguistic quality of the answer, including grammar, fluency, logical flow, coherence, and overall readability. \\[0.25em]
\addlinespace[0.5em]

\rowcolor{black!5}
\multicolumn{3}{c}{\textbf{\textcolor{blue}{Grounding \& Stability Metrics}}} \\
\addlinespace[0.4em]

\textit{Strict Faithfulness} & A, C &
Evaluates how many factual claims in the answer are directly supported by the retrieved context, enforcing strict evidence alignment (entity accuracy and temporal correctness) \\

\textit{Calibration} & Q, A, C &
Examines the stability of metric by measuring variance across different judge configurations (model choice and temperature). \\
\bottomrule
\end{tabular}
\caption{Summary of the \textsc{RAGVue} metrics.}
\label{tab:RAGVue-metrics-summary}
\end{table*}
\section{Streamlit UI Images} \label{appendix_sec:UI}

\begin{figure*}[ht]
  \centering
  \begin{subfigure}[b]{1.1\textwidth}
    \centering
    \includegraphics[width=\textwidth]{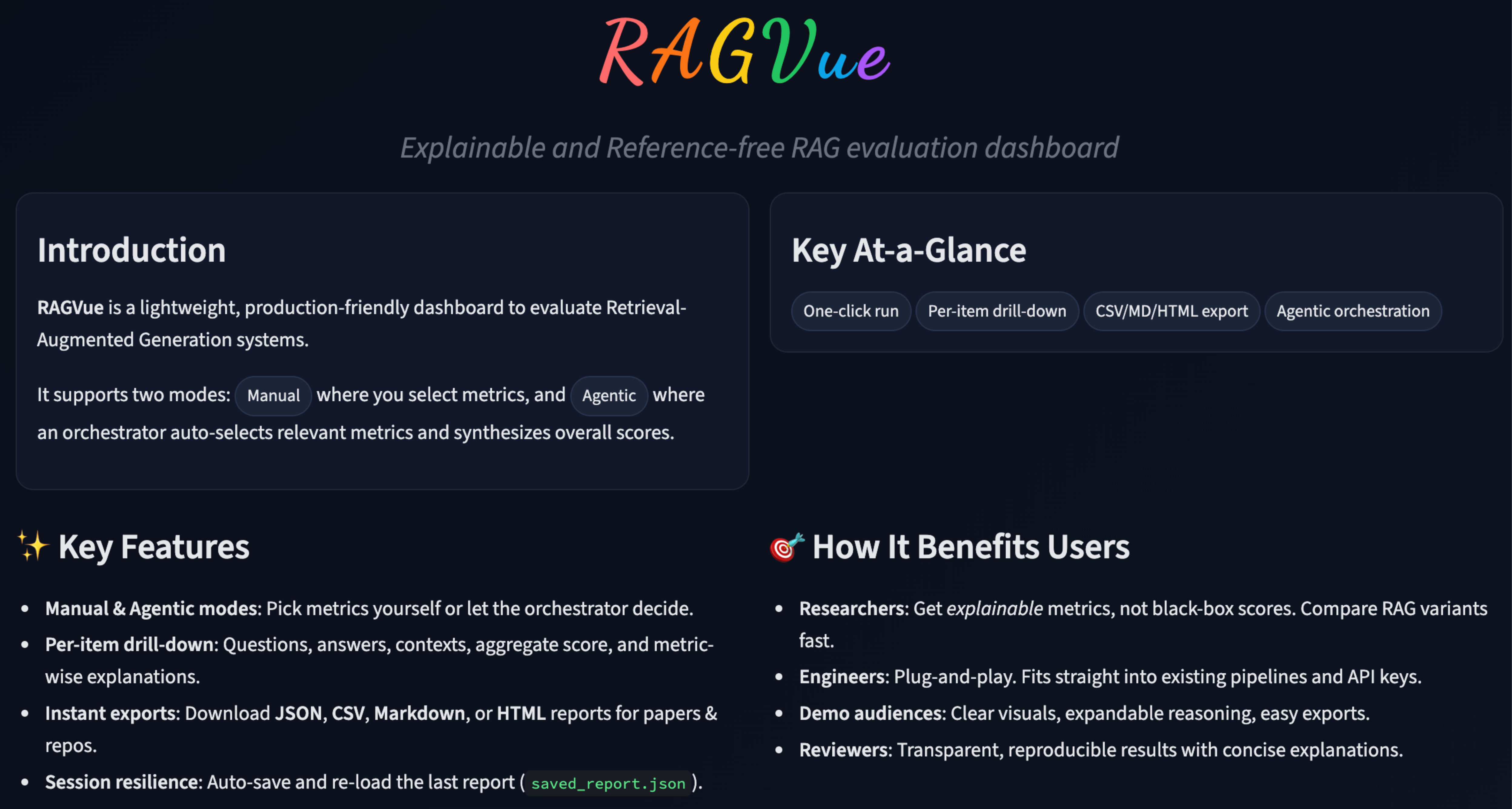}
    \caption{Introduction page.}
    \label{fig:mainpage}
  \end{subfigure}
  \hfill
  \begin{subfigure}[b]{1.1\textwidth}
    \centering
    \includegraphics[width=\textwidth]{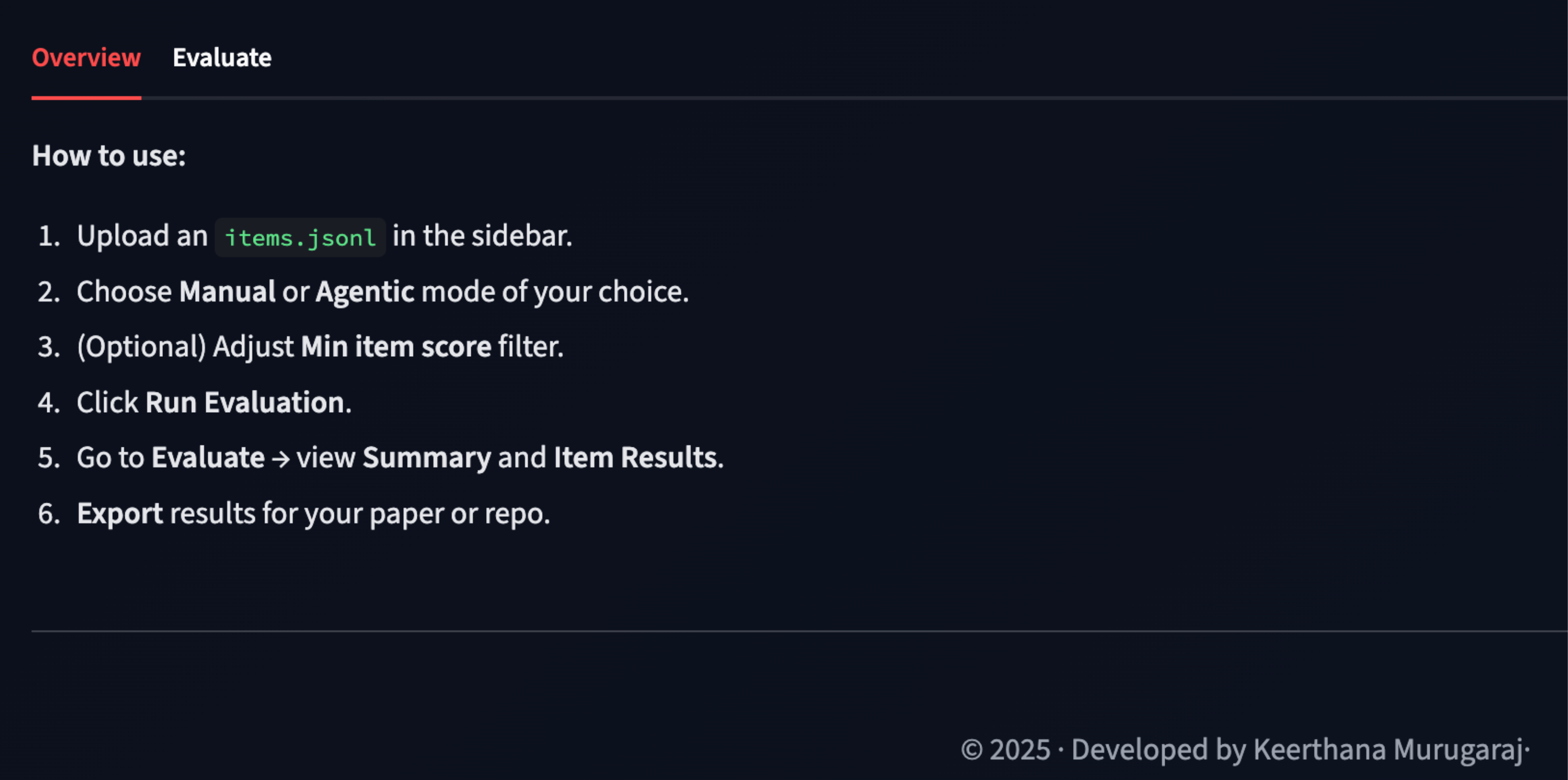}
    \caption{Overview tab: usage instructions.}
    \label{fig:usage}
  \end{subfigure} 
  \caption{\textsc{RAGVue} Streamlit UI: introduction page \& overview tab.}
  \label{fig:intro_usage}
\end{figure*}

\begin{figure*}[ht]
  \centering
  \begin{subfigure}[b]{0.48\textwidth}
    \centering
    \includegraphics[width=\textwidth]{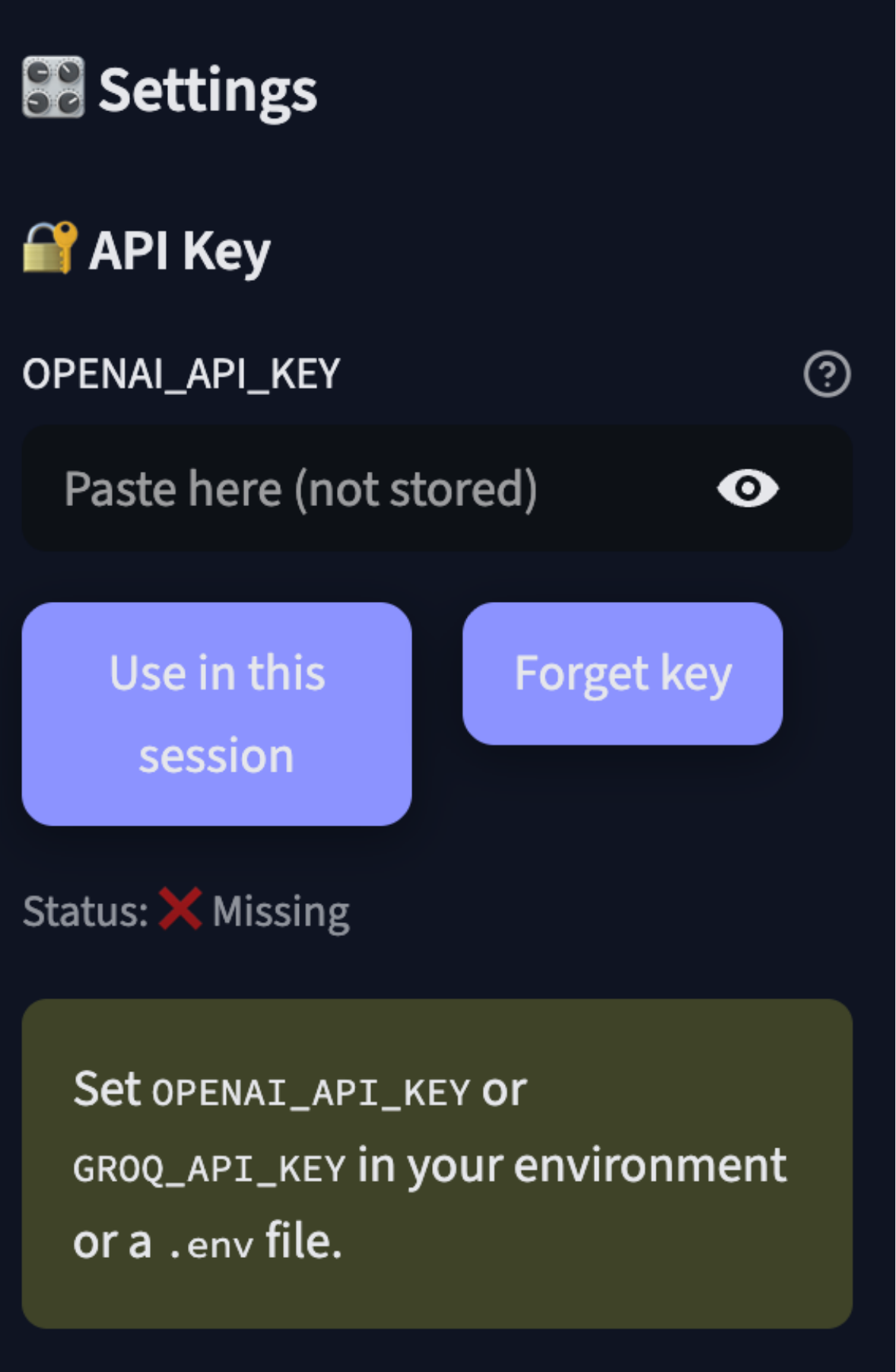}
    \caption{API and key settings.}
    \label{fig:ui-settings-api}
  \end{subfigure}
  \hfill
  \begin{subfigure}[b]{0.48\textwidth}
    \centering
    \includegraphics[width=\textwidth]{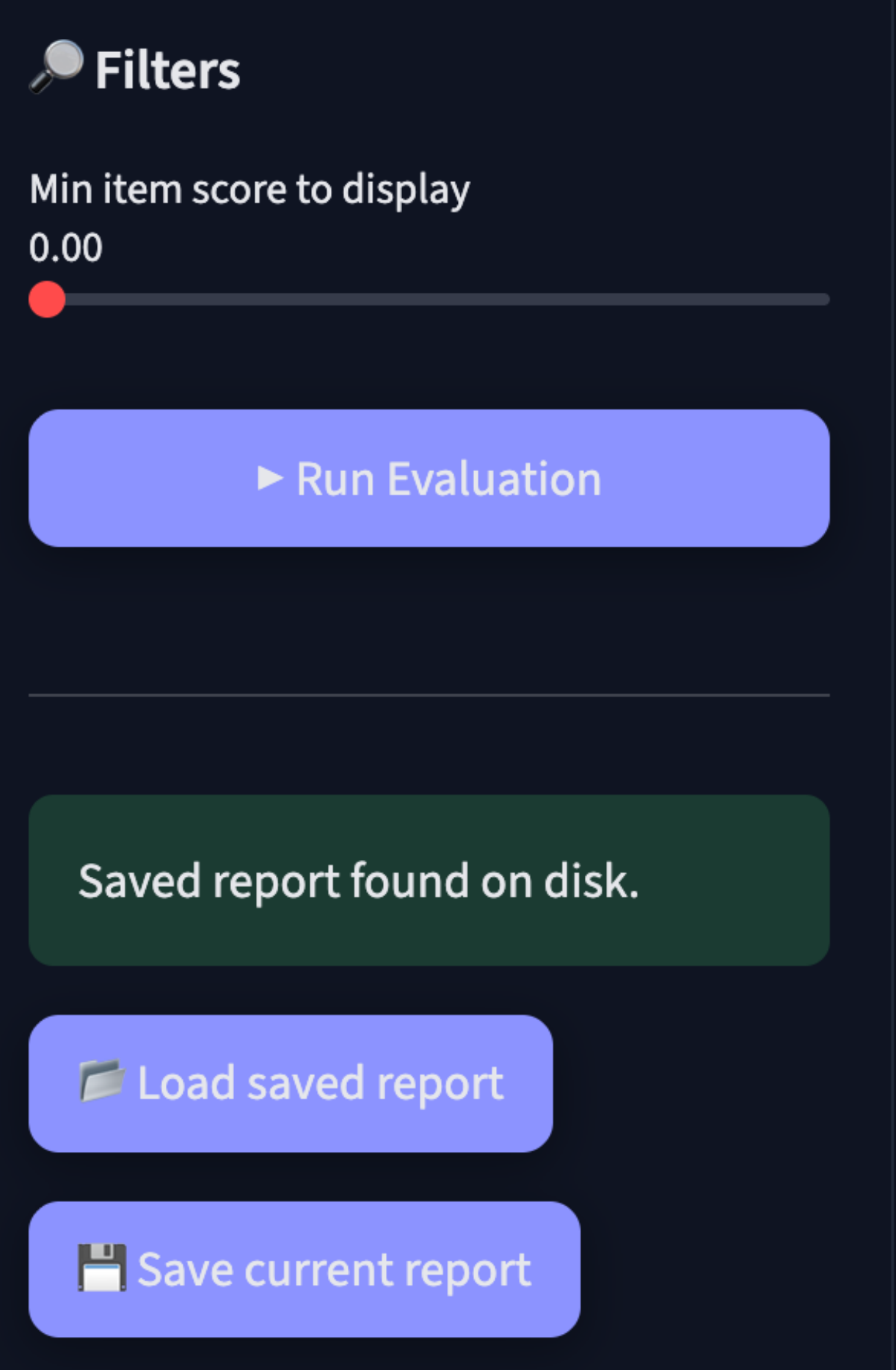}
    \caption{Filters and saving options.}
    \label{fig:ui-filters-save}
  \end{subfigure}

  \begin{subfigure}[b]{0.48\textwidth}
    \centering
    \includegraphics[width=\textwidth]{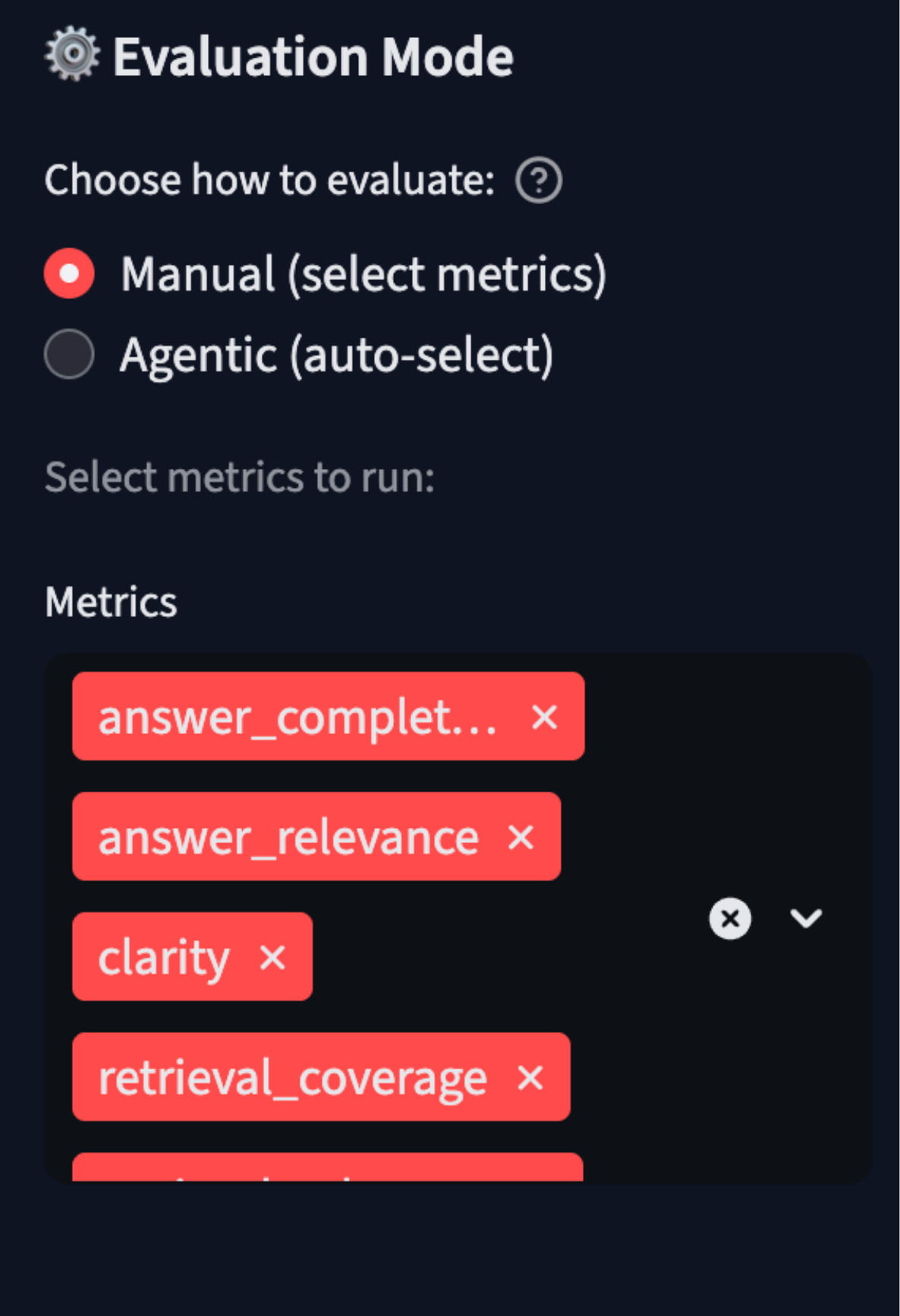}
    \caption{Manual/agentic mode configuration.}
    \label{fig:ui-manual-mode}
  \end{subfigure}
  \hfill
  \begin{subfigure}[b]{0.48\textwidth}
    \centering
    \includegraphics[width=\textwidth]{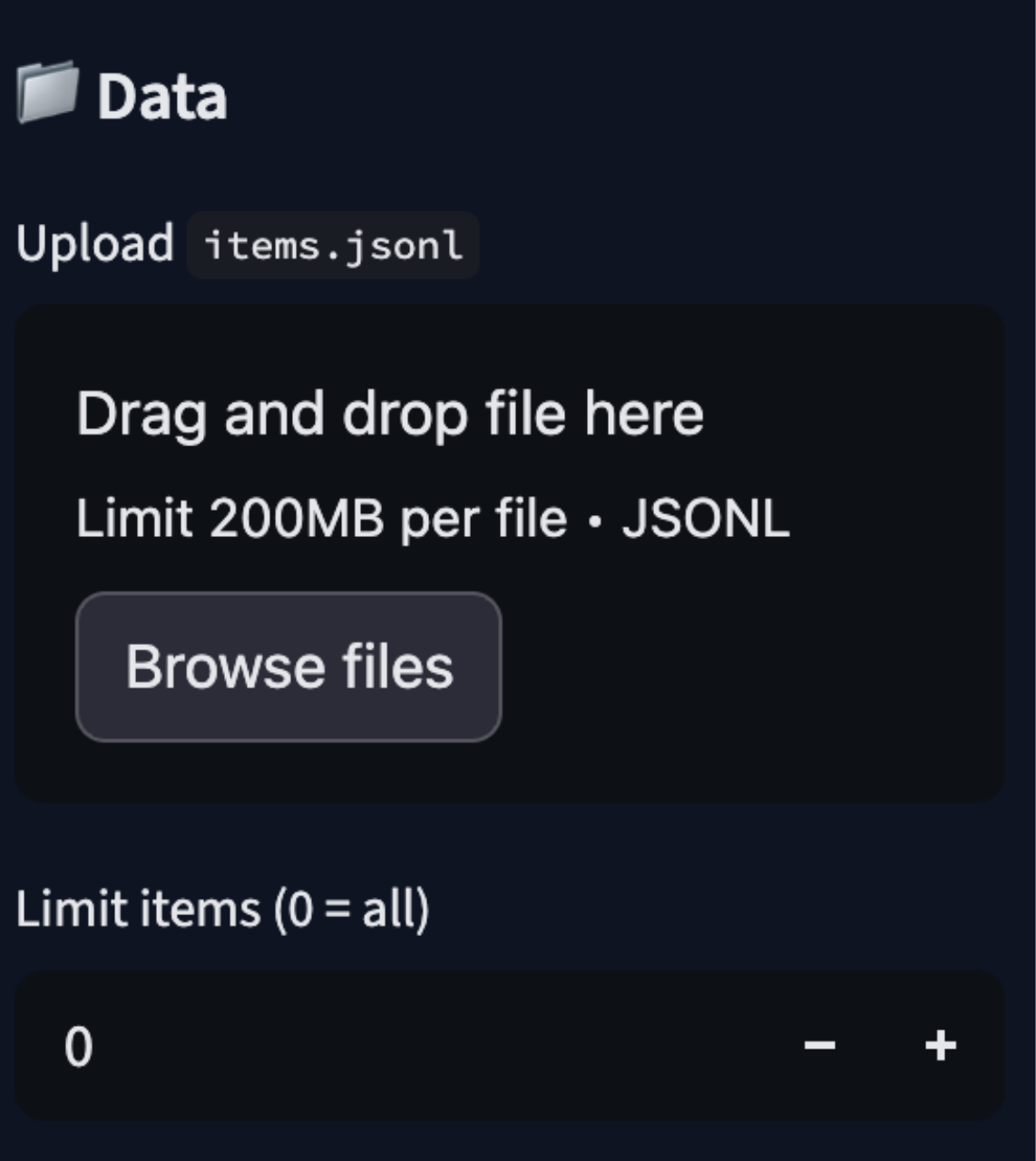}
    \caption{Data upload option.}
    \label{fig:ui-agentic-mode}
  \end{subfigure}

  \caption{\textsc{RAGVue} Streamlit user interface: API configuration, filter and save settings, manual/agentic mode selection, data upload options.}
  \label{fig:settings}
\end{figure*}

\begin{figure*}[ht]
  \centering

  \begin{subfigure}[b]{1.1\textwidth}
    \centering
    \includegraphics[width=\textwidth]{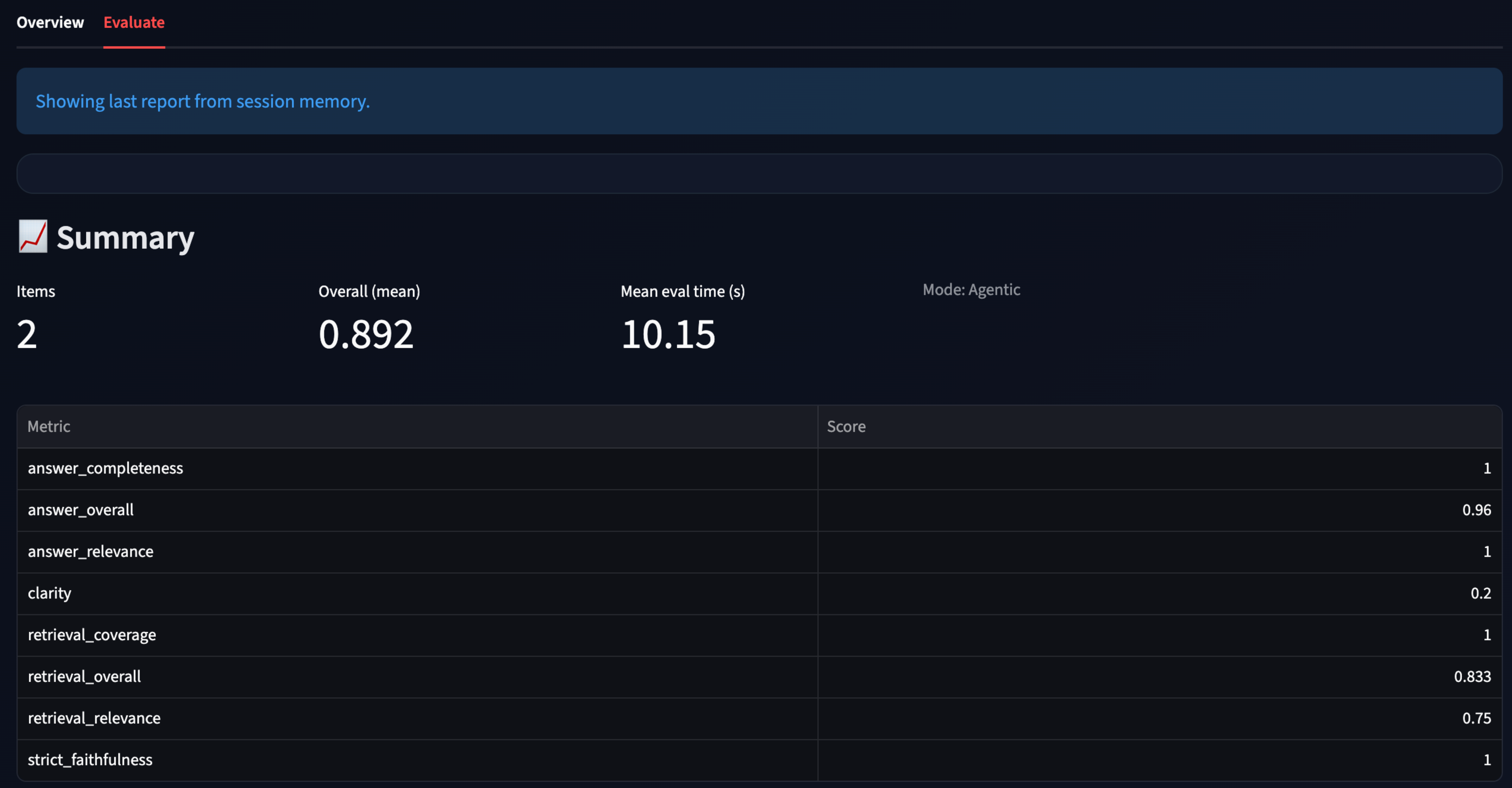}
    \caption{Evaluate tab showing the global summary across all cases.}
    \label{fig:evaluate-tab-summary}
  \end{subfigure}
  \hfill
  \begin{subfigure}[b]{1.1\textwidth}
    \centering
    \includegraphics[width=\textwidth]{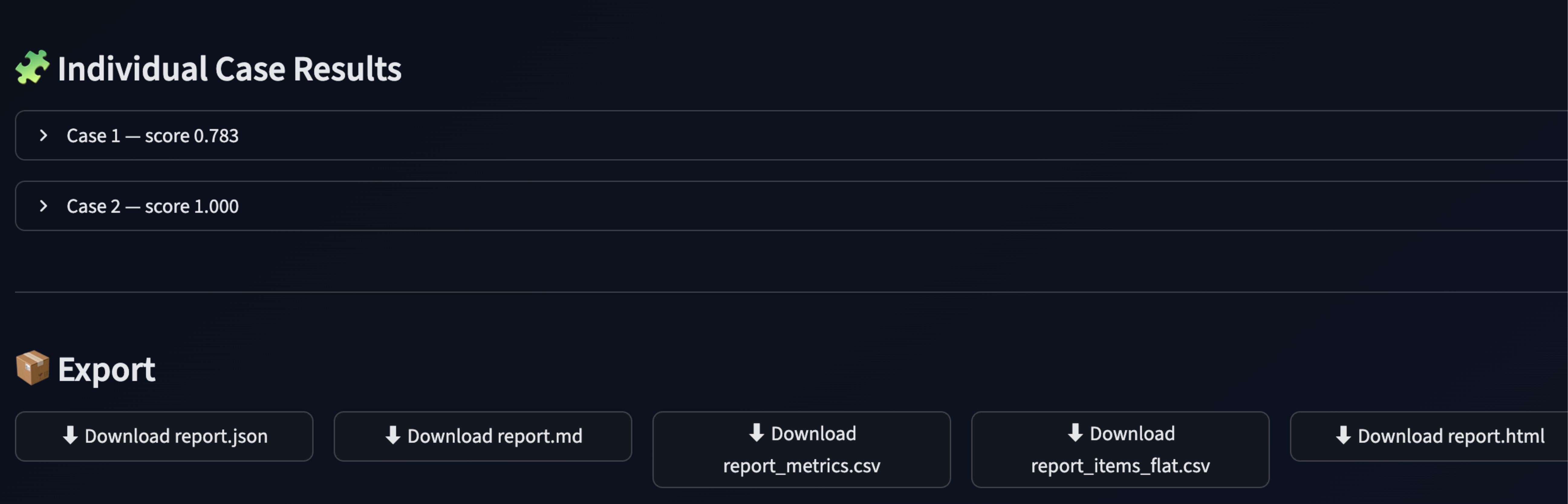}
    \caption{Detailed individual case report for a single (Q, A, C) instance.}
    \label{fig:individual-case-report}
  \end{subfigure}

  \caption{\textsc{RAGVue} Streamlit user interface: evaluation summary view and individual case report.}
  \label{fig:ui-evaluate}
\end{figure*}

\begin{figure*}[ht]
  \centering
  \begin{subfigure}[b]{1.1\textwidth}
    \centering
    \includegraphics[width=\textwidth]{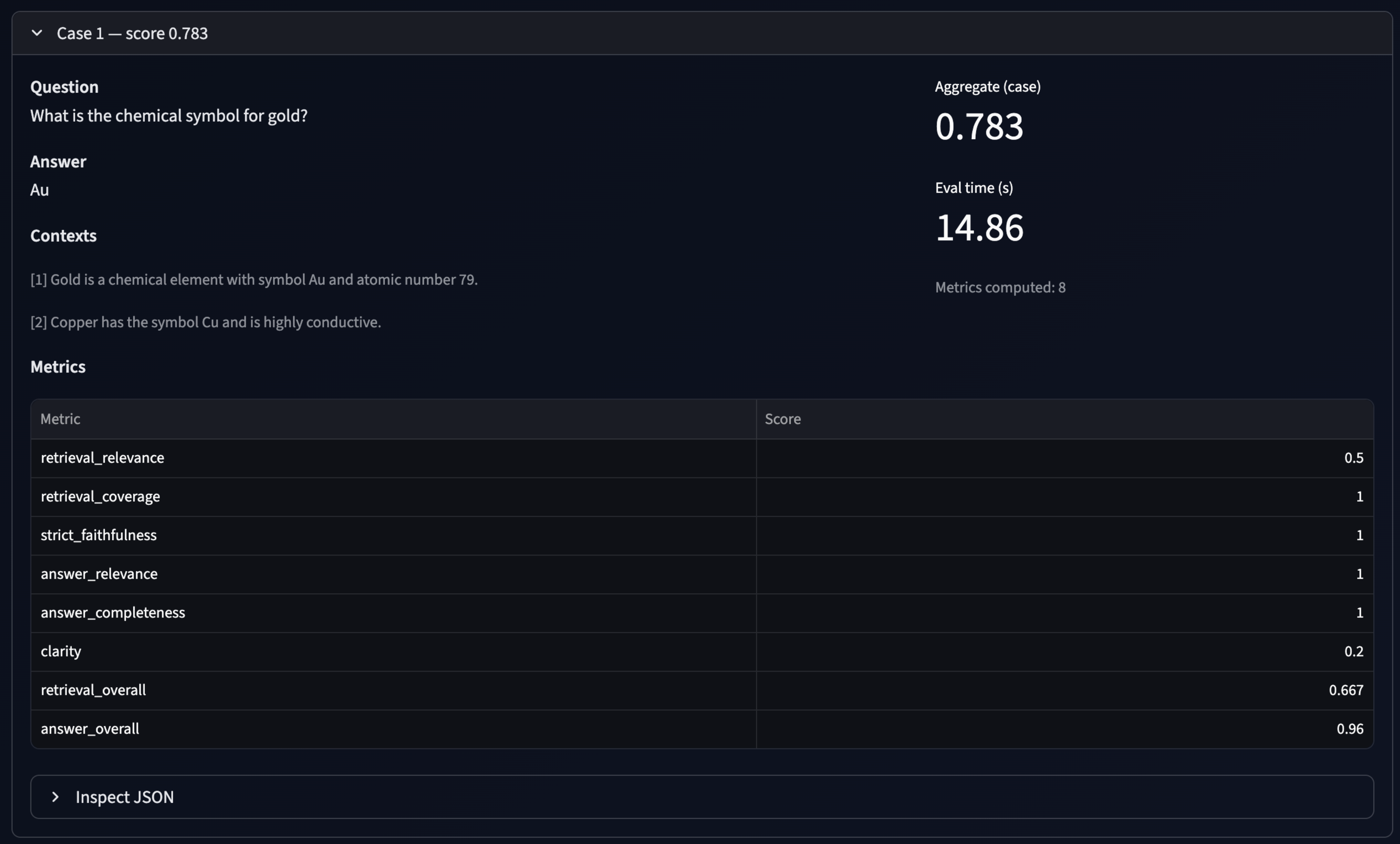}
    \caption{Agentic mode applied to (Q, A, C) triplets. 
             The orchestrator correctly selects all relevant metrics.}
    \label{fig:case1-agentic}
  \end{subfigure}
  \hfill
  \begin{subfigure}[b]{1.1\textwidth}
    \centering
    \includegraphics[width=\textwidth]{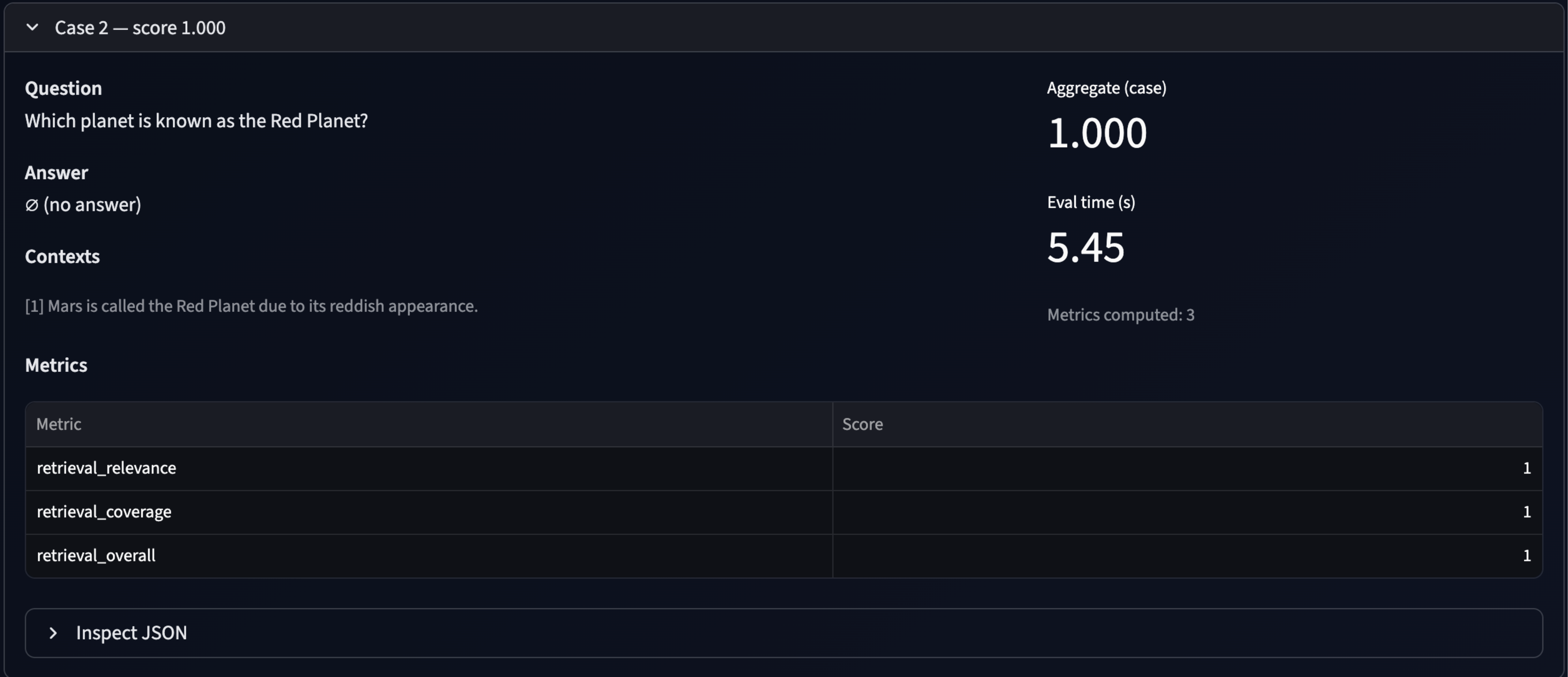}
    \caption{Agentic mode applied to (Q, C) triplets. 
             The orchestrator selects only retrieval metrics and skips answer metrics.}
    \label{fig:case2-agentic}
  \end{subfigure}
  \caption{Agentic mode behavior across different input configurations.}
  \label{fig:agentic-overview}
\end{figure*}

The following figures illustrate the full \textsc{RAGVue} Streamlit interface and its functionality.
Figure~\ref{fig:mainpage}-\ref{fig:usage} provides the introduction page and the overview tab, which guide users through the workflow and usage instructions.
Figure~\ref{fig:ui-settings-api}–\ref{fig:ui-agentic-mode} presents the core configuration components, including API key setup, data selection, manual and agentic mode configuration, and optional filters and report-saving tools.
Figure~\ref{fig:evaluate-tab-summary}-\ref{fig:individual-case-report} shows the evaluation tab, which contains both the global summary across all processed cases and the detailed report for an individual (Q, A, C) example.
Finally, Figure~\ref{fig:case1-agentic}-\ref{fig:case2-agentic} demonstrates the behavior of the agentic orchestrator for different input formats, highlighting its ability to select appropriate metrics based on the available fields.

\section{Limitations \& Future Work}

\textsc{RAGVue} represents our first step toward a transparent and diagnostic evaluation framework for Retrieval Augmented Generation. The current version delivers seven core metrics, two operational modes, and a no-code user interface, but there is still significant room for growth. As \textsc{RAGVue} relies on LLM-based evaluation, careful selection of the underlying judge models is recommended to ensure stable and consistent scoring. We plan to extend \textsc{RAGVue} with additional metrics, perform retrieval and grounding analysis on complex queries, and provide broader support for different LLM models. The agentic mode will become more adaptive, assisting users by detecting errors and automatically selecting appropriate metrics. Over time, our goal is to develop \textsc{RAGVue} into a unified evaluation pipeline with baseline models, system comparisons, and optional multimodal support.

\end{document}